\documentclass[10pt,twocolumn,letterpaper]{article}
\pdfoutput=1

\usepackage{cvpr}              

\usepackage{graphicx}
\usepackage[accsupp]{axessibility}
\usepackage{amsmath}
\usepackage{amssymb}
\usepackage{booktabs}
\usepackage{enumitem}
\usepackage{color}
\usepackage{xcolor}
\usepackage{textcomp}
\usepackage{epsfig}
\usepackage{multirow}
\usepackage{bigdelim}
\usepackage{wrapfig}
\usepackage{caption}
\usepackage{subcaption}

\usepackage[pagebackref,breaklinks,colorlinks]{hyperref}

\definecolor{ColorLightBlue}{RGB}{18, 137, 255}

\newcommand{\ra}[1]{\renewcommand{\arraystretch}{#1}}

\newlength\secmargin
\newlength\paramargin
\newlength\figmargin

\newcommand{\mypara}[1]{\vspace{2mm}\noindent\textbf{#1}~}

\newcommand{\accuchange}[1]{$_{\color{teal} (#1)}$}

\def\OurMethodName{VADER}

\usepackage[capitalize]{cleveref}
\crefname{section}{Sec.}{Secs.}
\Crefname{section}{Section}{Sections}
\Crefname{table}{Table}{Tables}
\crefname{table}{Tab.}{Tabs.}

\def\cvprPaperID{1769} 
\def\confName{CVPR}
\def\confYear{2023}

\usepackage{textcomp}
\usepackage{float}
\usepackage{dblfloatfix} 
\usepackage{multicol}

\begin{document}

\title{Generalizable Local Feature Pre-training for Deformable Shape Analysis}

\author{ Souhaib Attaiki \hspace{1.5cm} Lei Li \hspace{1.5cm} Maks Ovsjanikov\\
LIX, \'Ecole Polytechnique, IP Paris}

\maketitle

\begin{abstract}
Transfer learning is fundamental for addressing problems in settings with little training data. While several transfer learning approaches have been proposed in 3D, unfortunately, these solutions typically operate on an entire 3D object or even scene-level and thus, as we show, fail to generalize to new classes, such as deformable organic shapes. In addition, there is currently a lack of understanding of what makes pre-trained features transferable across significantly different 3D shape categories. In this paper, we make a step toward addressing these challenges. First, we analyze the link between feature locality and transferability in tasks involving deformable 3D objects, while also comparing different backbones and losses for local feature pre-training. We observe that with proper training, learned features \emph{can be} useful in such tasks, but, crucially, only with an appropriate choice of the receptive field size. We then propose a differentiable method for optimizing the receptive field within 3D transfer learning. Jointly, this leads to the first learnable features that can successfully generalize to unseen classes of 3D shapes such as humans and animals. Our extensive experiments show that this approach leads to state-of-the-art results on several downstream tasks such as segmentation, shape correspondence, and classification. Our code is available at  \url{https://github.com/pvnieo/vader}.
\end{abstract}	
	

\section{Introduction}
\label{sec:intro}

Extracting informative representations from 3D geometry is a central task in Computer Vision, Computer Graphics, and related fields. Classical approaches have relied on hand-crafted features derived from basic geometric principles \cite{johnson1999using,belongie2000shape,pottmann2009integral,sun2009concise,aubry2011wave}.
More recently, the focus has shifted towards data-driven approaches that learn features directly from 3D data  \cite{guo2020deep,cao2020comprehensive,bronstein2017geometric} in a task-specific manner.

In addition to methods that learn features from scratch for each application, several recent works have also advocated for general-purpose \textit{representation learning} on geometric data \cite{xie2020pointcontrast,hou2020exploring,wang2021unsupervised}. Inspired by the success of transfer learning in other domains \cite{zhuang2020comprehensive}, these methods aim to learn informative representations of 3D data, which can then be exploited in data-limited downstream tasks. 

\begin{figure}
    \centering
    \includegraphics[width=\columnwidth]{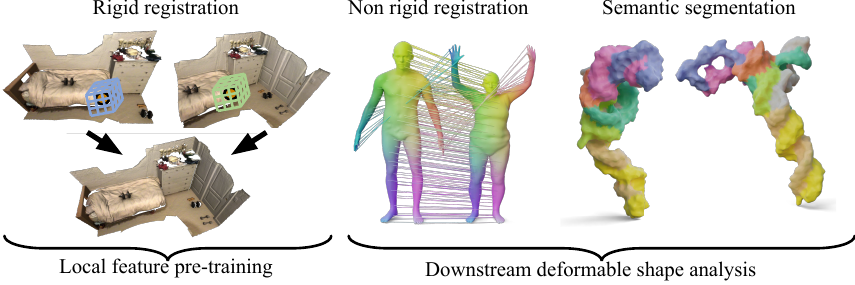}
    \caption{We present \OurMethodName{}, a novel feature pre-training technique aiming for  deformable shapes. By pre-training \emph{local} feature extractors on 3D scenes for rigid alignment, our approach enables transfer learning to downstream deformable shape analysis tasks, such as shape matching and semantic segmentation.}
    \label{fig:teaser}
    \vspace{\figmargin}
\end{figure}

Despite this progress, state-of-the-art architectures in \emph{deformable shape analysis} still either rely on classical hand-crafted features as input signals to their learning pipelines \cite{monti2017geometric,poulenard2018multi,litany2017deep,sharp2020diffusionnet}, or are trained \textit{from scratch} for each task \cite{groueix20183d,donati2020deep,li2020shape}, thus requiring significant amounts of labeled data. Unfortunately, as we demonstrate in our work, existing 3D representation learning approaches fail to provide a useful signal in tasks that involve highly deformable shapes, such as  shape correspondence or segmentation.

This result is perhaps expected since existing approaches have primarily focused on transfer learning across man-made 3D objects or scenes \cite{xiao2022unsupervised}, and are typically restricted to settings with significant domain overlap between training and test data. Furthermore, there is currently a lack of understanding of what makes pre-trained features transferable, especially across significantly different shape classes.



\begin{figure*}[t]
    \centering
    \includegraphics[width=\textwidth]{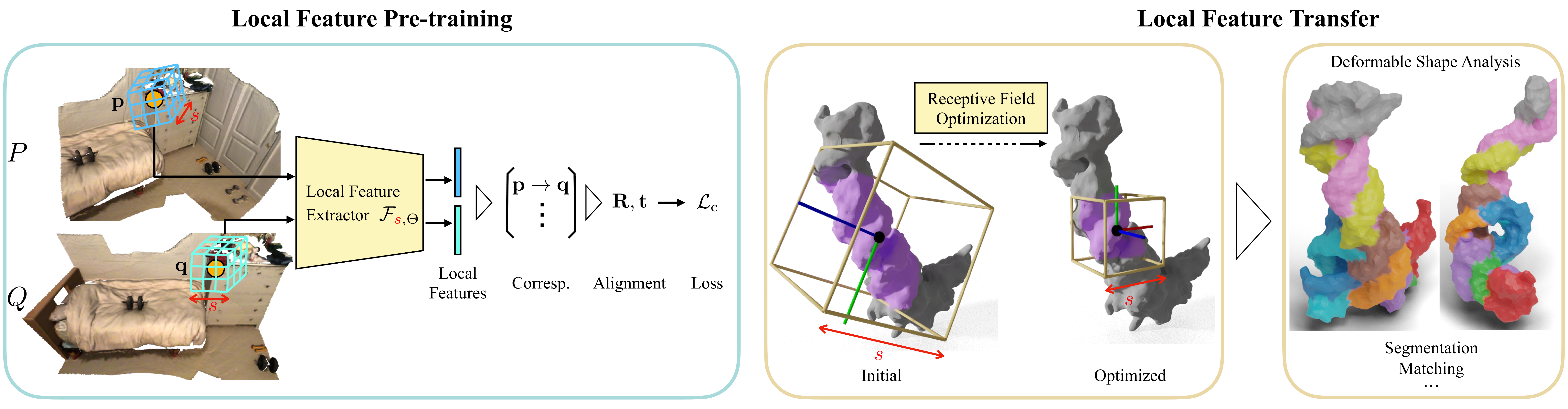}
    \caption{\textbf{Method overview}. We propose generalizable \textit{local feature} pre-training for deformable shape analysis. We first pre-train a local feature extractor $\mathcal{F}_{s, \Theta}$, which has a learnable receptive field size $s$ and network parameters $\Theta$, on a pretext task of matching local features for 3D alignment. We then propose a differentiable method for optimizing the receptive field size $s$ to transfer $\mathcal{F}_{s, \Theta}$ to downstream tasks. For illustration purposes, we use a molecular surface segmentation task as an example on the right.}
    \label{fig:pipeline}
    \vspace{\secmargin}
\end{figure*}

In this work, we aim to investigate the transferability of geometric features to develop representation learning approaches that are useful in downstream deformable shape analysis tasks, such as non-rigid shape matching and semantic segmentation (see \cref{fig:teaser}). Taking inspiration from recent studies that emphasize the importance of low and mid-level features in enabling 2D transfer learning \cite{matsoukas2022makes,zhao2021what}, we explore the impact of feature locality on downstream task accuracy across significantly different 3D shape categories. Our study shows that, with a carefully chosen architecture, successful general-purpose representation learning for deformable 3D shape analysis is possible. We also find that the receptive field (or local support) size plays a crucial role in the transferability of features and needs to be adapted between training and test data. To address this, we propose a receptive field optimization strategy, which, combined with a specific pre-training approach, leads to state-of-the-art results on a wide range of downstream tasks. An overview of our proposed method can be found in \cref{fig:pipeline}.

To summarize, our main contributions are as follows:
\begin{enumerate}[topsep=3pt,partopsep=3pt,itemsep=3pt,parsep=0pt]
    \item We investigate the link between the locality of geometric (3D) features and their transferability in challenging deformable shape tasks.
    \item We build upon the investigation and propose a novel method for optimizing the receptive field size of local features in the context of transfer learning in 3D tasks. We demonstrate that this optimization brings significant improvement and allows pre-trainable features to generalize well to unseen data in downstream tasks.
    \item We show that pre-training local features with an unsupervised cycle consistency loss outperforms the standard contrastive PointInfoNCE loss.
    \item Based on all of these insights, we propose a new local feature pre-training mechanism and show its utility in a wide range of tasks involving deformable objects, going beyond man-made objects or scenes considered in previous 3D transfer learning approaches.
\end{enumerate}

\section{Related Work}
Extracting robust local features is a key problem in 3D shape analysis, and a full review is beyond the scope of this paper. Below we discuss methods most closely related to ours and refer the readers to both early \cite{heider2011local,bronstein2012feature,guo20143d} and more recent surveys including \cite{guo2016comprehensive,guo2020deep,cao2020comprehensive,bronstein2017geometric} for a comprehensive overview of local feature extraction and geometric deep learning more broadly.

\mypara{Hand-crafted features.}
Early efforts in designing informative features for 3D geometry have focused primarily on either ensuring invariance to rigid motion \cite{johnson1999using,kortgen20033d,tombari2010unique,zaharescu2009surface,knopp2010hough} or intrinsic features invariant to isometries.
Intrinsic descriptors are typically based on either analysis of geodesic distances \cite{hilaga2001topology,gal2007pose} or derived quantities such as spectral properties arising from the eigenbases of the Laplace-Beltrami operator \cite{sun2009concise,aubry2011wave,bronstein2010scale}.
Intrinsic descriptors, such as the HKS \cite{sun2009concise} and WKS \cite{aubry2011wave}, are
a very popular choice in deformable shape correspondence methods, especially based on the functional maps framework, e.g., \cite{ovsjanikov2012functional,aflalo2013spectral,huang2014functional,eynard2016coupled,burghard2017embedding,rodola2017partial,ren2018continuous}, among many others.

\mypara{Learning for deformable shape matching.}
To overcome the limitations of hand-crafted features, more recent approaches have tried to learn descriptors directly from deformable shape data \cite{litman2013learning,masci2015geodesic,attaiki2023clover,boscaini2016learning,poulenard2018multi,wiersma2020cnns}.
Remarkably, however, many learning-based works still use hand-crafted features (most commonly, SHOT, HKS, WKS, or similar) as input to their learning pipelines \cite{litany2017deep,roufosse2019unsupervised,lim2018simple,maron2017convolutional,wang2020mgcn,eisenberger2020deep,sharp2020diffusionnet}.
Several recent works \cite{groueix20183d,marin22_why,fey2018splinecnn,li2020shape,donati2020deep,attaiki2022ncp,attaiki2021dpfm} investigate learning robust deformable shape correspondence directly from raw geometry, either by exploiting extensive training sets or combining spatial and spectral regularization \cite{donati2020deep,eisenberger2020deep}. Nevertheless, the features learned in these works are typically application and dataset-specific and fail to generalize to new shape classes and shape processing tasks.

\mypara{Learning for man-made shape matching.}
A parallel line of studies has focused on learning local geometric features for man-made object or scene alignment.
Many efforts have been made to explore different representations for local 3D geometry \cite{zeng20173dmatch,gojcic2019perfect,khoury2017learning,deng2018ppfnet,huang2017learning,li2020end}.
With the advancement of 3D deep learning techniques, networks based on PointNet \cite{wang2019deep,wang2019prnet,yew20183dfeat,aoki2019pointnetlk,yew2020rpm}, sparse convolution \cite{choy2019fully,choy20194d}, or kernel point convolution \cite{bai2020d3feat} have been applied to dense feature extraction.
However, these networks have very limited generalization ability even across 3D scene datasets, such as from indoor scenes to outdoor scans \cite{bai2020d3feat}, due to local features being coupled with global scene structures.

\mypara{Feature pretraining.} 
Most closely related to ours are the recent PointContrast \cite{xie2020pointcontrast} and its follow-up works \cite{hou2020exploring,chen20214dcontrast,wang2021unsupervised} that explore the potential of learning informative representations for 3D data, which can then be leveraged in downstream tasks. In contrast to these works, which focus on man-made objects or scenes, we consider generalizable feature pre-training for \textit{deformable shape analysis}, and study generalizability across significantly different 3D shape categories. Most importantly, our work highlights the impact of feature locality on transferability, which is lacking in prior works.

\section{Motivation}
\label{sec:motivation}

Our main objective is to build a general-purpose feature extractor that can be applied to highly deformable shape analysis tasks, such as matching human shapes or segmentation of molecular surfaces, among others.
We first examine existing designs of geometric feature pre-training and perform a pilot study to understand their generalization power in such  tasks.
Our key insight is that the \textit{locality} of geometric features plays a crucial role in their transferability across different categories, which has so far been overlooked in prior works. To this end, we first perform an in-depth analysis of feature locality versus transferability in a representative deformable shape matching task. 

\mypara{Revisiting PointContrast.}
PointContrast \cite{xie2020pointcontrast} is a recent feature pre-training framework, in which a geometric feature extractor is pre-trained by a pretext task involving correspondences on 3D scenes \cite{dai2017scannet} related by rigid motion.
Specifically, PointContrast uses a fully-convolutional sparse U-Net \cite{choy2019fully} to output a feature vector for every point in the input.
During pre-training, given two scene fragments, a contrastive PointInfoNCE loss is used to minimize the feature distance for corresponding points and maximize it for non-corresponding ones:
\begin{equation}
    \label{eq:NCELoss}
    \mathcal{L}_{\text{nce}} = - \sum_{(i,j) \in \Omega} \log \dfrac{\exp(\mathbf{f}_i \cdot \mathbf{f}_j / \tau)}{\sum_{(\cdot, k) \in \Omega} \exp(\mathbf{f}_i \cdot \mathbf{f}_k / \tau)},
\end{equation}
where $\Omega$ is the set of corresponding point pairs in the overlap region, $\mathbf{f}_{\square}$ denotes the learned point-wise feature vector, and $\tau$ is a temperature hyper-parameter.

\mypara{Pilot study on deformable shapes.}
We study the link between the locality of geometric features and their transferability through the lens of a downstream deformable shape matching task \cite{roufosse2019unsupervised,halimi2019unsupervised,eisenberger2020deep,donati2020deep,sharma2020weakly}.
Specifically, we leverage a widely-used human shape dataset, FAUST-Remeshed (FR) \cite{Ren2019,Bogo2014}, consisting of 100 humans in diverse poses.
Given a pair of 3D shapes as input, we first use pre-trained feature extractors to compute a feature vector for every point in the input.
We then find correspondences via nearest neighbor search between the extracted features and apply a lightweight refinement with ZoomOut \cite{Melzi_2019}, a common practice in prior works.
We use the standard mean geodesic error \cite{Kim2011} as the evaluation metric.
Note that there is no fine-tuning of network weights, and thus this study provides a good indication of how informative and transferable the pre-trained features are in downstream applications.

\begin{figure}
    \centering
    \includegraphics[width=1.0\columnwidth]{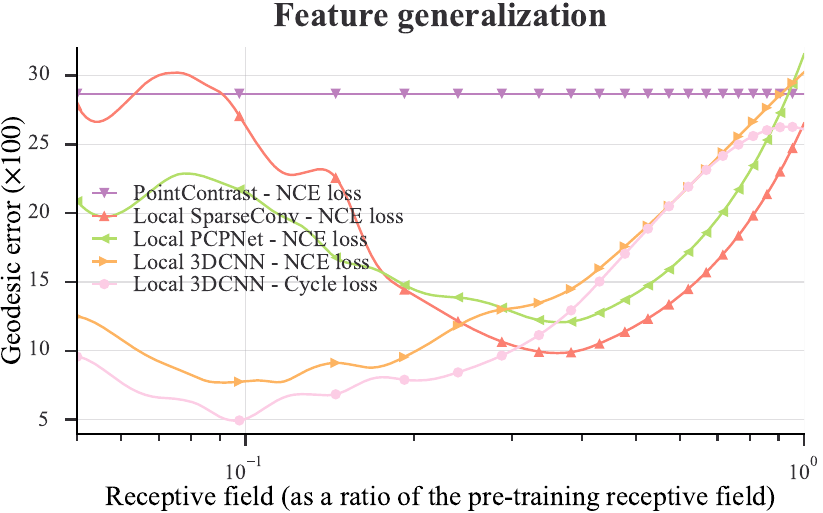}
    \caption{Feature locality vs. transferability in a downstream task of non-rigid shape matching on the FAUST-Remeshed dataset. \emph{Local} in the legend denotes local patch-based input at each point.}
    \label{fig:locality_vs_transferability}
\end{figure}

\mypara{Feature locality vs. transferability.}
We evaluated the features produced by PointContrast in this context and obtained a matching error of \textbf{28.7}, compared to \textbf{6.1} achieved by a recent axiomatic method \cite{ren2018continuous}.  
%
We attribute this limited utility of PointContrast features for deformable shapes to the global structure of its network, which employs a fully-convolutional U-Net design. Furthermore, this network is trained on entire 3D scenes with a global receptive field, making it significantly less likely to generalize to unseen shape categories.

To address this issue, we propose to limit the receptive field size and pre-train feature extractors \textit{that take as input only a local patch centered at each point}, and output a feature vector for the center point (\cref{fig:pipeline}). Intuitively, the space of local patches is significantly smaller than the space of shapes, thus potentially enabling generalization across different shape categories \cite{surfacerec07,patchnet2020,chabra2020,ao2021spinnet,pcpnet2018,fujiwara2011locally}.

To evaluate this general approach, we select three different architectures for pre-training a local feature extractor:
a) SparseConv \cite{choy20194dminkski,xie2020pointcontrast}, a sparse tensor-based network, which also constitutes the backbone of PointContrast;
b) PCPNet \cite{pcpnet2018,qi2017pointnet}, a PointNet-based architecture; 
and c) 3D CNN \cite{gojcic2019perfect,li2021updesc}, operating on voxel grids.
We then pre-train these local feature extractors for \textit{the rigid alignment task} on 3D man-made scenes \cite{zeng20173dmatch}, following a similar strategy as in PointContrast. We follow the standard design choices and optimal pre-training patch size as used in the existing literature. Please see the exact architectures and pre-training details in the supplementary.

Given these pre-trained local feature extractors, a natural question would be how to adapt the receptive field (patch) size between pre-training and downstream 3D data, which may consist of significantly different shape classes, to make the local feature extractors generalize well. For this, we test a wide range of receptive field sizes (as ratios of the pre-training one) and plot their corresponding matching performance on FR in \cref{fig:locality_vs_transferability}.

When comparing PointContrast and Local SparseConv in \cref{fig:locality_vs_transferability}, we observe that making the network \textit{local} and operating on patches significantly improves feature transferability, especially for some specific receptive field size in this downstream task. Moreover, we observe that Local 3D CNN has the best generalization performance on FR, compared to Local SparseConv and Local PCPNet.

Most importantly, this pilot study highlights the importance of feature locality and the crucial role that the optimal receptive field size plays in the downstream task for successful transfer learning.
In practice, performing an exhaustive search of receptive field sizes is typically infeasible. To address this issue, we propose a differentiable approach to feature locality optimization for downstream 3D geometric data (\cref{subsec:local_feature_transfer}). 

\section{Methodology}
\label{sec:method}

In this section, we first briefly introduce our chosen local feature pre-training method in \cref{subsec:local_feature_pretrain}.
We then describe our novel differentiable method for optimizing the receptive field size of local features for transfer learning across datasets in \cref{subsec:local_feature_transfer}.

In the following, we represent our local feature extractor as $\mathcal{F}_{s, \Theta}$, where $s \in \mathbb{R}$ is a \emph{learnable} receptive field size (\cref{sec:motivation}) and $\Theta$ denotes the parameters of the network backbone.
Given a shape $P$ represented as either a triangle mesh or a point cloud and a point $\mathbf{p} \in P$, the feature extractor $\mathcal{F}_{s, \Theta}$ associates a feature vector to $\mathbf{p}$, which summarizes the local 3D geometry around $\mathbf{p}$.
The feature extractor is composed of two stages:
local patch extraction, denoted by $\mathcal{G}_{s}(\mathbf{p}, P)$,
and geometric feature extraction, denoted by $\mathcal{H}_{\Theta}(\mathcal{G}_{s}(\mathbf{p}, P)) = \mathcal{F}_{s, \Theta} (\mathbf{p}, P)$.

\subsection{Local Feature Pre-training}
\label{subsec:local_feature_pretrain}

\begin{figure}[t!]
    \centering
    \includegraphics[width=0.9\columnwidth]{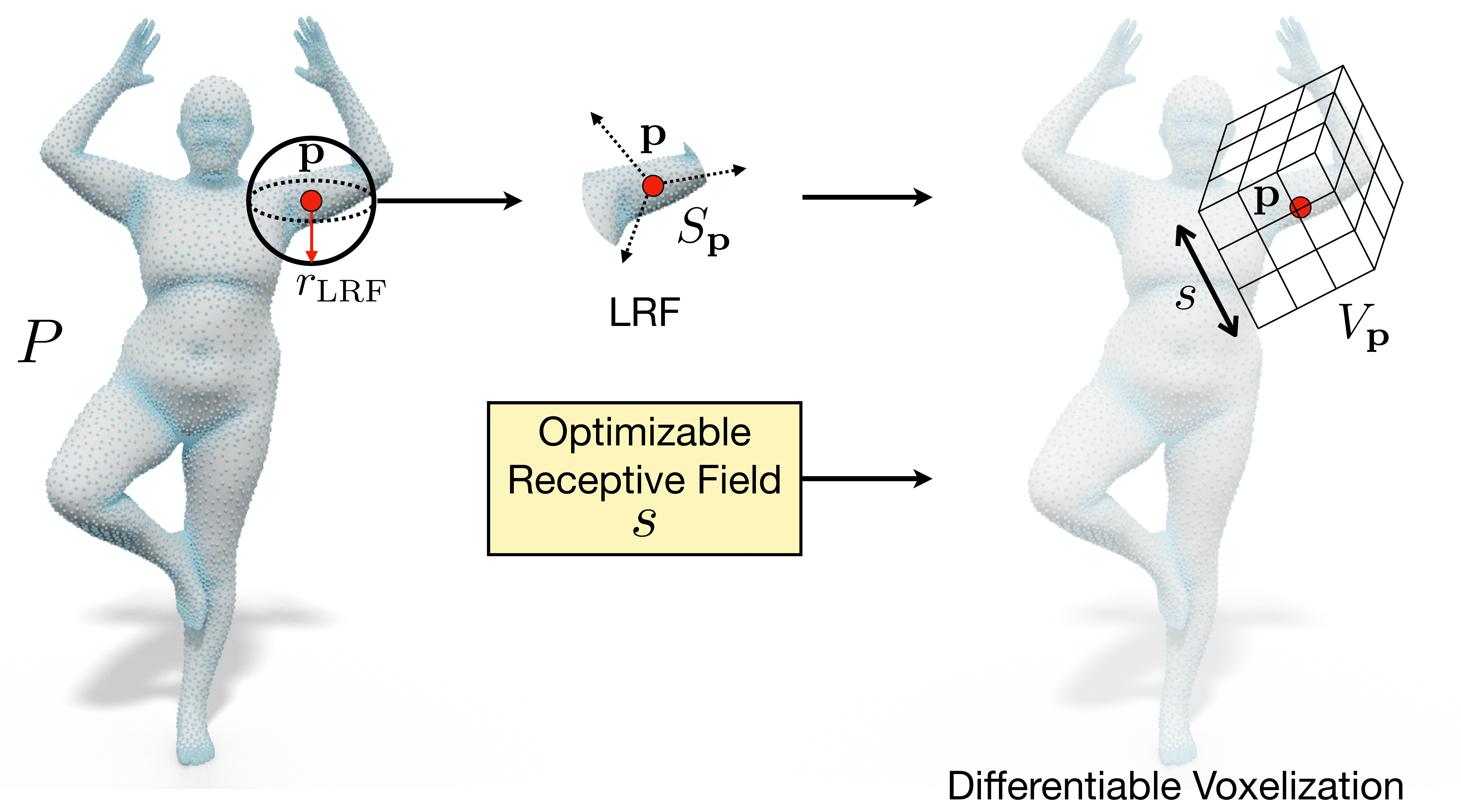}
    \caption{To extract local features, we parameterize the 3D geometry around point $\mathbf{p} \in P$ with a voxel-based representation $V_\mathbf{p}$, reoriented by a local reference frame (LRF) and endowed with an optimizable receptive field size $s$.}
    \label{fig:input_parameterization}
    \vspace{\figmargin}
\end{figure}

\mypara{Feature extraction.}
Following the transferability study in \cref{fig:locality_vs_transferability}, we build $\mathcal{F}_{s, \Theta}$ by adapting the local feature extractor architecture from WSDesc \cite{li2021updesc}, which is based on a voxel-based representation \cite{zeng20173dmatch,gojcic2019perfect} and a 3D CNN for geometric learning.
\cref{fig:input_parameterization} illustrates the local patch extraction procedure.
A voxel grid $V_{\mathbf{p}}=\mathcal{G}_{s}(\mathbf{p}, P)$ is anchored at $\mathbf{p}$, whose size is determined by the optimizable $s$, to capture the local geometric structure.
$V_{\mathbf{p}}$ is reoriented by a local reference frame (LRF) for rotation invariance in local features \cite{gojcic2019perfect,li2021updesc}.
The voxel values in $V_{\mathbf{p}}$ are computed by a differentiable voxelization layer based on probabilistic aggregation \cite{li2021updesc} for back-propagating gradients to $s$.
Next, for geometric feature extraction, $V_{\mathbf{p}}$ is fed to a 3D CNN $\mathcal{H}_{\Theta}$, which consists of six convolutional layers with ReLU and normalization in-between \cite{gojcic2019perfect}.
The resulting feature vector $\mathcal{H}_{\Theta}(V_{\mathbf{p}})$ is 32-dimensional.

\mypara{Pre-training loss.}
We adopt the pretext task of matching local features for 3D scene alignment, as advocated in PointContrast \cite{xie2020pointcontrast}, to pre-train the feature extractor $\mathcal{F}_{s, \Theta}$.
This allows us to gain access to large-scale 3D training data, which has been absent thus far for deformable shape analysis.
In this work, we examine two different loss functions for pre-training: one is the PointInfoNCE loss $\mathcal{L}_{\text{nce}}$ (\cref{sec:motivation}), and the other is an unsupervised loss $\mathcal{L}_{\text{c}}$ based on cycle consistency, introduced in \cite{li2021updesc} and briefly described below for completeness.

Cycle consistency evaluates the extracted features in a rigid registration pipeline without using ground truth labels.
Given two shapes $P$ and $Q$ and their per-point features extracted by $\mathcal{F}_{s, \Theta}$, putative correspondences between $P$ and $Q$ are first constructed by differentiable nearest neighbor search \cite{plotz2018neural} in the learned feature space.
Then a 3D transformation aligning $P$ to $Q$ is estimated from the correspondences and represented as a rotation $\mathbf{R} \in \mathbb{R}^{3 \times 3}$ and a translation $\mathbf{t} \in \mathbb{R}^{3}$.
Let $\mathbf{R}'$ and $\mathbf{t}'$ denote the computed alignment, in the reverse direction, from $Q$ to $P$.
The cycle consistency loss enforces the composition of $(\mathbf{R}, \mathbf{t})$ and $(\mathbf{R}', \mathbf{t}')$ to be an identity matrix as follows:
\begin{equation}
    \label{eq:CycleConsistencyLoss}
    \mathcal{L}_{\text{c}} = \| \mathbf{R}\mathbf{R}' - \mathbf{I} \|_1 + \| \mathbf{R}\mathbf{t}' +  \mathbf{t}\|_1,
\end{equation}
where $\mathbf{I}$ is an identity matrix.
We refer the reader to \cite{li2021updesc} for more details.

As shown in \cref{fig:locality_vs_transferability}, pre-training with $\mathcal{L}_{\text{c}}$ leads to local features having significantly better transferability on FR, compared to pre-training with $\mathcal{L}_{\text{nce}}$.

This is remarkable as the PointInfoNCE loss is a go-to choice for representation learning across many domains and application scenarios \cite{Feyetal2020,chen20214dcontrast,li2022srfeat}. We ascribe this performance difference partly to the fact the PointInfoNCE loss focuses on whether \textit{individual} point correspondences are correct or not, and does not consider local or global feature consistency \cite{li2022srfeat}.
In contrast, the cycle consistency loss jointly assesses all the extracted local features and thus imposes correspondence consistency between $P$ and $Q$ for successful global alignment. As a result, $\mathcal{L}_{\text{c}}$ promotes more consistent (and smoother) local features to be learned for nearby points than $\mathcal{L}_{\text{nce}}$.

To validate this hypothesis, we measure the Dirichlet energy \cite{pinkall1993computing} of the pre-trained features (the lower, the smoother) on the human shapes of FR, resulting in 86.8 for $\mathcal{L}_{\text{nce}}$ and 75.4 for $\mathcal{L}_{\text{c}}$.
These measurements confirm that the local features pre-trained by the cycle consistency loss are smoother (and thus more consistent) than those by the PointInfoNCE loss, which can potentially benefit downstream applications, especially across different datasets. 

\subsection{Local Feature Transfer}
\label{subsec:local_feature_transfer}

As we demonstrated in \cref{sec:motivation}, the size of the receptive field is of crucial importance for downstream tasks. In this section, we introduce a receptive field optimization method, which automatically determines the best receptive field for each downstream dataset.
To this end, we take advantage of the full differentiability of our network design and optimize the learnable receptive field size $s$, so that the distribution of features extracted on the target dataset is similar to that of features extracted by the pre-trained network on the pre-training dataset.

We draw inspiration from the domain adaptation literature \cite{kundu_cvpr_2020,liu2021sourcefree,liang2020really,huang2021model}, where the main goal is to transfer knowledge from a network trained in a domain with abundant labeled data to a domain of interest with minimal or unlabeled data. This is achieved by minimizing the \textit{domain shift} between the feature distribution of the two domains using some distance in the probability measure space. Thus we propose to  minimize the discrepancy between the distribution of features extracted from the pre-training dataset, and the distribution of features extracted from the target (downstream) dataset. For this, we base our approach on the Maximum Mean Discrepancy (MMD) used in previous works \cite{liu2021sourcefree,qin2019pointdan,6751384,Borgwardt2006IntegratingSB}.
Intuitively, networks are pre-trained to have a strong response signal under the distribution of training local patches, and thus our goal is to optimize the local patch sizes, given \textit{unlabeled} downstream data, so that the extracted features have a similar distribution.

In practice, after pre-training the feature extractor, we sample $n_s$ points from the pre-training (source) dataset and extract the features from these points with the pre-trained network. We denote these features as $F^s = \{f_i^s\}_{i=1..n_s}$.
For each new downstream (target) dataset, we formulate an optimization problem, where we freeze the network backbone weights $\Theta$ but optimize the receptive field size $s$, so that the network produces features as close to $F^s$ as possible, using the MMD metric. Our network is fully differentiable, and we solve this problem via gradient descent to find the optimal receptive field $s$. This optimized receptive field is then used for all downstream tasks involving this dataset. At each iteration during optimization, we randomly sample $n_t$ points from the target dataset and compute features for them with the pre-trained network. We denote these features as $\{f_i^t\}_{i=1..n_t}$ and formulate the loss as follows:
\begin{align}
	\nonumber
E_{mmd} =& \frac{1}{n_s n_s} \sum_{i, j=1}^{n_s} \kappa(f_i^s, f_j^s) + \frac{1}{n_s n_t} \sum_{i, j=1}^{n_s, n_t} \kappa(f_i^s, f_j^t) \\
&+ \frac{1}{n_t n_t} \sum_{i, j=1}^{n_t} \kappa(f_i^t, f_j^t),
	\label{eq:loss_mmd}
\end{align}
where $\kappa$ is the Radial Basis Function (RBF) kernel.
The Adam optimizer \cite{kingma2017adam} is used in gradient descent.
 

\begin{figure}[h!]
    \centering
    \includegraphics[width=\linewidth]{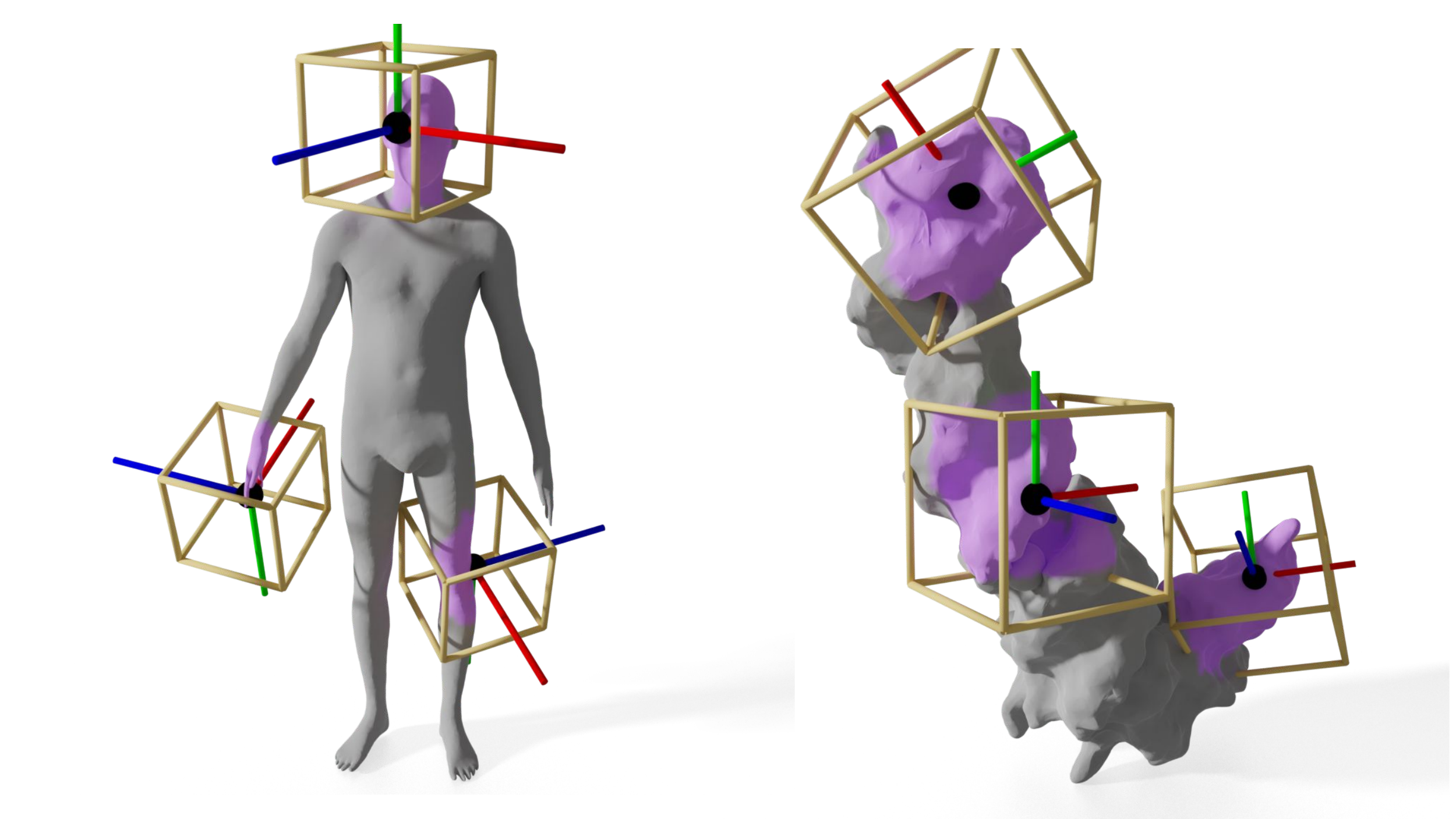}
    \caption{Qualitative visualization of the receptive field of local patches used in our downstream experiments on both the human and RNA datasets. The points of interest are indicated in black. The surface areas in pink represent the local receptive fields \emph{after  optimization}. We also show the corresponding voxel grids and the XYZ axes of the local reference frames.}
    \label{fig:patch_size}
    \vspace{\figmargin}
\end{figure}

\section{Experiments}
\label{sec:experiments}

In this section, we provide extensive experiments to highlight the generalization power of our local feature pre-training and receptive field size optimization methods in a suite of downstream shape analysis tasks especially involving highly deformable, organic shapes.
We consider diverse benchmarks including human and partial animal shape matching as well as molecular surface segmentation.
We reuse the same pre-trained local feature extractor $\mathcal{F}_{s, \Theta}$ and then perform our receptive field size optimization once individually on each deformable shape dataset (\cref{subsec:local_feature_transfer}).

\mypara{Implementation.}
We denote our features as \OurMethodName{} for Voxelized Alignment-based DEscriptoR.
In the pre-training stage, we use the 3DMatch dataset \cite{zeng20173dmatch} and train the local feature extractor $\mathcal{F}_{s, \Theta}$ for 16K steps.
We use the Adam optimizer with a learning rate of $10^{-3}$  for network weight update.
For receptive field size optimization, we use the same learning rate for Adam, and we take $n_s = 10^4$  extracted patches from the pre-training dataset. 

\mypara{Baselines.}
We compare a wide spectrum of hand-crafted and pre-trained features in deformable shape tasks.
For the hand-crafted features, we consider the Heat Kernel Signature (HKS)  \cite{sun2009concise}, Wave Kernel Signature (WKS) \cite{aubry2011wave}, and SHOT descriptors \cite{tombari2010unique}, as well as the straightforward vertex positions (XYZ).
For the pre-trained features, we use the PointContrast features learned with the PointInfoNCE loss (PCN) or a hardest-contrastive loss (PCH) \cite{xie2020pointcontrast}.


\subsection{Human Shape Matching}
\label{subsec:human_matching}

\begin{table}[!t]
    \begin{center}
    \ra{1.0}
        \resizebox{0.81\columnwidth}{!}{%
            \begin{tabular}{@{} lrr @{} }
                \toprule
                \textbf{Method / Dataset}                & \textbf{FR}-\textbf{SH} & \textbf{SR}-\textbf{SH} \\
                \midrule
                SURFMNET \cite{roufosse2019unsupervised} & 30.1                    & 28.6                    \\
                Cyclic FMaps \cite{ginzburg2019cyclic}   & 36.5                    & 38.6                    \\
                WSupFMNet \cite{sharma2020weakly}        & 26.3                    & 30.2                    \\
                Deep Shells \cite{eisenberger2020deep}   & 26.3                    & 22.8                    \\
                \midrule
                DiffusionNet - XYZ                       & 22.4                    & 23.3                    \\ 
                DiffusionNet - HKS                       & 10.4                    & 15.4                    \\ 
                DiffusionNet - WKS                       & 9.3                     & 24.0                    \\ 
                DiffusionNet - SHOT                      & 10.8                    & 21.5                    \\ 
                DiffusionNet - PCH                       & 25.8                    & 33.2                    \\ 
                DiffusionNet - PCN                       & 22.6                    & 36.2                    \\ 
                DiffusionNet - \OurMethodName{} (ours)   & \textbf{6.4}            & \textbf{6.9}            \\ 
                \bottomrule
            \end{tabular}
        }
        \caption{Performance of various features for unsupervised deformable shape matching on un-aligned data. X-Y means training on X and testing on Y. Values are mean geodesic error $\times 100$ on unit-area shapes.}
        \label{tab:unaligned_unsup}
    \end{center}
    \vspace{\figmargin}
\end{table}

\mypara{Unsupervised matching.}
We perform unsupervised shape matching \cite{roufosse2019unsupervised,eisenberger2020deep} on the FAUST-Remeshed (FR), SCAPE-Remeshed (SR), and SHREC'19 datasets (SH) \cite{Ren2019,Bogo2014,Anguelov2005,shrec19}, consisting of 100, 71, and 44 \emph{unaligned} human shapes in different poses, respectively. 
The same train/test splits in prior works \cite{sharma2020weakly,eisenberger2020deep} are adopted.
We feed the above baseline features and our \OurMethodName{} respectively as input to a surface learning backbone DiffusionNet \cite{sharp2020diffusionnet}, which produces a functional map \cite{ovsjanikov2012functional,litany2017deep} as output for a given pair of shapes.
We leverage the unsupervised functional map losses \cite{sharma2020weakly} for training the backbone.
The evaluation metric is the mean geodesic error of predicted maps with respect to the ground truth on unit-area shapes \cite{Kim2011}. We use X-Y to denote training on dataset X and testing on dataset Y.

\begin{figure}
    \centering
    \includegraphics[width=\columnwidth]{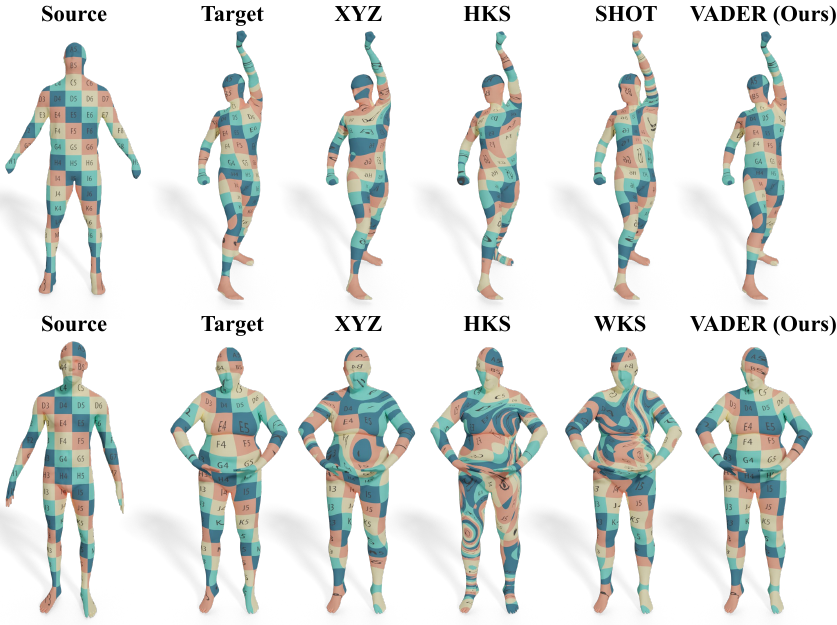}
    \caption{Qualitative comparisons of human shape matching by texture transfer. Top: results of unsupervised matching on SH. Bottom: results of supervised matching on FQ. The best three performing competitors are shown.}
    \label{fig:qual_all_human}
    \vspace{\figmargin}
\end{figure}

As shown in \cref{tab:unaligned_unsup}, our approach yields the best results for unsupervised shape correspondence in both FR-SH and SR-SH settings, while the other tested features fail to achieve reasonable matching performance.
We stress that this is a challenging test case as most existing unsupervised methods rely on aligned shapes, e.g.,  \cite{sharma2020weakly,eisenberger2021neuromorph}.
Note that the PointContrast features perform worse than the hand-crafted features, indicating the overfitting to the pre-training data distribution, as discussed in \cref{sec:motivation}, and thus the limited transferability.
Our approach also outperforms several recent unsupervised approaches in \cref{tab:unaligned_unsup}, including SURFMNET \cite{roufosse2019unsupervised}, Cyclic FMaps \cite{ginzburg2019cyclic}, WSupFMNet \cite{sharma2020weakly}, and Deep Shells \cite{eisenberger2020deep}. 
The comparisons clearly demonstrate the utility of incorporating generalizable pre-trained features, which is missing in prior works. 
We provide qualitative comparisons in \cref{fig:qual_all_human} (Top), showing that only our approach leads to visually plausible results.

\mypara{Robustness to meshing.}
We evaluate the performance of supervised shape matching \cite{donati2020deep} on the original FAUST (FO) dataset \cite{Bogo2014} and its remeshed version by quadratic error simplification (FQ) \cite{sharp2020diffusionnet} to demonstrate the robustness and generalization of our approach against significant mesh connectivity changes across datasets. 
We build a point-wise MLP network \cite{litany2017deep} (to reduce the dependence on the backbone architecture) on top of the baseline features and our \OurMethodName{} respectively, and then predict functional maps for shape pairs.
The MLP backbones are trained on FO with predictions supervised by the ground-truth maps with a simple $L_2$ loss.
We report the mean geodesic error metric on FQ.

\begin{figure}
    \centering
    \includegraphics[width=\columnwidth]{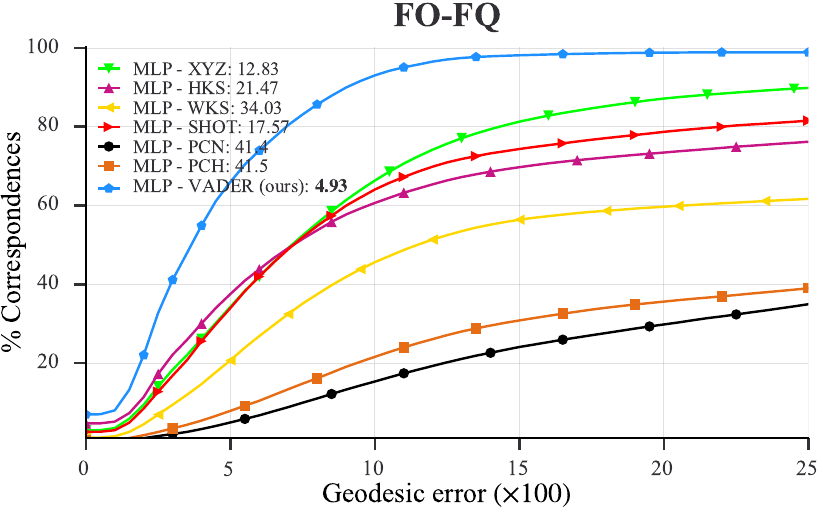}
    \caption{Accuracy of various features for supervised shape matching when the connectivity changes from training to test (mean errors $\times 100$ are reported in the legend).}
    \label{fig:super_fo_fq}
\end{figure}

\cref{fig:super_fo_fq} shows the correspondence quality with a varying error threshold. 
It can be seen that hand-crafted features such as SHOT degrade rapidly under remeshing. 
Although WKS and HKS are intrinsic features and do not depend on the meshing connectivity, they are not expressive enough by themselves and need to be combined with a powerful network backbone such as DiffusionNet, instead of the point-wise MLPs used in this experiment.
Differently, our approach achieves superior generalization performance in this challenging setting, showing that \OurMethodName{} is highly robust to remeshing and resampling, and effectively captures local geometric structures in deformable shapes.
\cref{fig:qual_all_human} (bottom) presents a qualitative comparison of the computed maps with the three best-performing competitors.

\subsection{Molecular Surface Segmentation}
\label{subsec:molecular_segmentation}

\begin{table}[!t]
     \begin{center}
    \ra{1.0}
          \resizebox{\columnwidth}{!}{%
               \begin{tabular}{@{}lrrr@{}}
                    \cmidrule[\heavyrulewidth]{1-4}
                    \textbf{Method}                                 & \multicolumn{3}{c}{\textbf{Accuracy $\pm$ s.d}}                                                           \\
                    \cmidrule{2-4}
                                                                    & Full Dataset                                    & 50 Shapes                  & 100 Shapes                 \\
                    \cmidrule[\heavyrulewidth]{1-4}
                    PointNet++ \cite{qi2017pointnet}                & 74.4\%                                          & --                         & --                         \\
                    PCNN \cite{atzmon2018pcnn}                      & 78.0\%                                          & --                         & --                         \\
                    SPHNet \cite{poulenard2019effective}            & 80.1\%                                          & --                         & --                         \\
                    SplineCNN \cite{fey2018splinecnn}               & 53.6\%                                          & --                         & --                         \\
                    SurfaceNetworks \cite{kostrikov2018surface}     & 88.5\%                                          & --                         & --                         \\

                    \cmidrule[\heavyrulewidth]{1-4}
                    DiffusionNet - XYZ \cite{sharp2020diffusionnet} & 90.5 $\pm$ 0.6\%                                & 82.7 $\pm$ 0.63\%          & 83.4 $\pm$ 0.67\%          \\
                    DiffusionNet - HKS                              & 90.6 $\pm$ 0.15\%                               & 82.7 $\pm$ 0.16\%          & 84.5 $\pm$ 0.09\%          \\
                    DiffusionNet - WKS                              & 88.7 $\pm$ 0.26\%                               & 77.6 $\pm$ 0.16\%          & 81.2 $\pm$ 0.20\%          \\
                    DiffusionNet - SHOT                             & 92.1 $\pm$ 0.08\%                               & 81.6 $\pm$ 0.31\%          & 85.7 $\pm$ 0.11\%          \\
                    DiffusionNet - PCH                              & 90.3 $\pm$ 0.1\%                                & 79.9 $\pm$ 0.59\%          & 83.6 $\pm$ 0.08\%          \\
                    DiffusionNet - PCN                              & 90.1 $\pm$ 0.09\%                               & 80.1 $\pm$ 0.29\%          & 83.4 $\pm$ 0.28\%          \\
                    DiffusionNet - \OurMethodName{} (ours)          & \textbf{92.6 $\pm$ 0.02\%}                      & \textbf{83.2 $\pm$ 0.20\%} & \textbf{86.8 $\pm$ 0.09\%} \\ 
                    \cmidrule[\heavyrulewidth]{1-4}
               \end{tabular}
          }
          \caption{Accuracy of various mesh and point cloud based methods for RNA segmentation. The reported numbers are mean accuracy over 5 runs randomly initialized. $\pm$ denotes standard deviation.}
          \label{tab:rna_seg}
     \end{center}
    \vspace{\figmargin}
\end{table}

Next, we conduct experiments in the molecular surface segmentation task, which aims to segment RNA molecules into functional components.
We use the dataset introduced in \cite{poulenard2019effective}, consisting of 640 RNA triangle meshes, where
each vertex is labeled into one of 259 atomic categories.
The dataset has an 80/20\% split for training and test sets.
We feed the baseline features and our \OurMethodName{} respectively as input to DiffusionNet and train it to predict a label at each vertex as output.

As shown in \cref{tab:rna_seg}, our approach achieves state-of-the-art segmentation performance when used with the full training set, outperforming both hand-crafted and pre-trained PointContrast features as well as several recent shape segmentation networks, such as \cite{kostrikov2018surface}.
We also perform experiments in more challenging settings, where only a fraction of the training set, with respectively 50 and 100 shapes (corresponding to 9\% and 18\% of the training set), is used.
We observe in \cref{tab:rna_seg} that our method consistently outperforms the competitors by a significant margin when given limited training data.
The results highlight that our pre-training and receptive field size optimization strategies bring significant improvement to downstream organic shape analysis tasks.

\subsection{Partial Animal Matching}
\label{subsec:animal_matching}
We also evaluate how well different geometric features perform on deformable shapes in the presence of significant partiality.
For this, we test on the challenging SHREC16' Cuts dataset \cite{cosmo2016shrec}, where the animal classes (cat, centaur, dog, horse, and wolf) are used for partial shape matching.
We follow the setup of DiffusionNet described in \cref{subsec:human_matching} for correspondence prediction. 

\begin{figure}[!t]
    \centering
    \includegraphics[width=\columnwidth]{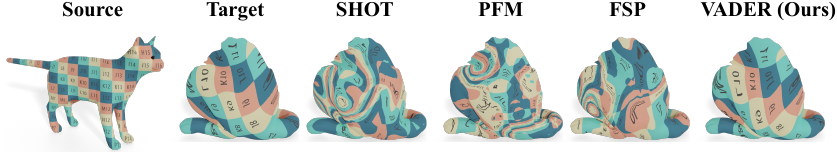}
    \caption{Qualitative comparisons of partial animal matching by texture transfer on the cat class of the SHREC'16 CUTS Animals benchmark.}
    \label{fig:qual_sup_animal}
    \vspace{\figmargin}
\end{figure}

We compare our approach to XYZ and SHOT as they are widely used in partial matching pipelines, in addition to full-fledged methods PFM \cite{Rodol2016} and FSP \cite{Litany2017}, specifically tailored toward partial shape matching. 
The results are summarized in \cref{tab:sup_animal}. Qualitative results are visualized in \cref{fig:qual_sup_animal}. Observe that our approach outperforms the competitors in this setting by a significant margin, including specially-tailored partial matching methods PFM and FSP.

\begin{table}[!t]
    \begin{center}
    \ra{1.0}
        \resizebox{0.9\columnwidth}{!}{%
        \begin{tabular}{@{} lr @{}}
            \toprule
            \textbf{Method / Dataset}              & SHREC'16 CUTS Animals \\
            \midrule
            PFM \cite{Rodol2016}                   & 8.8                   \\
            FSP \cite{Litany2017}                  & 12.2                  \\
            DiffusionNet - XYZ                     & 4.9                   \\
            DiffusionNet - SHOT                    & 4.6                   \\ 
            DiffusionNet - \OurMethodName{} (ours) & \textbf{3.7}          \\
            \bottomrule
        \end{tabular}
        }
        \caption{Performance (mean geodesic error $\times 100$) of various features on the SHREC'16 CUTS Animals benchmark.\vspace{-2mm}}
        \label{tab:sup_animal}
    \end{center}
    \vspace{\figmargin}
\end{table}

\subsection{Shape Classification}
\label{subsec:shape_classification}



We use the ShapeNet dataset \cite{chang2015shapenet} to demonstrate the effectiveness of our pre-trained feature extractor on man-made objects. We follow the classification setup from PointContrast \cite{xie2020pointcontrast}, using pre-trained weights as initialization for fine-tuning a classification network. Here our feature extractor is pre-trained with the PointInfoNCE loss. We conduct experiments on the ShapeNetCore v2 dataset with the same train/test split as PointContrast. We also consider a limited fine-tuning data setup, using a fraction of the fine-tuning data (1\% or 10\%). The classification accuracy comparison is summarized in \cref{table-shapenet_classification_accuracy_retrain}. It can be seen that feature pre-training improves performance compared to training from scratch. Also, our network achieves higher classification accuracy than PointContrast in all training setups.


\begin{table}[!t]
    \begin{center}
        \resizebox{0.99\columnwidth}{!}{%
        \ra{1.0}
        \begin{tabular}{@{} lrrrr @{}}
            \toprule
                \% train data    &  \multicolumn{2}{c}{\em{PointContrast}} & \multicolumn{2}{c}{\em{\OurMethodName{} (ours)}}   \\
                & From scratch  & PointInfoNCE & From scratch  & PointInfoNCE \\
                \midrule
                1\% data & 53.2 & 60.7 & 59.5 & \textbf{66.5}\\
                10\% data & 74.4  & 73.7 & 72.2 & \textbf{77.2}\\
                100\% data & 76.9 & 77.2 & 79.0 & \textbf{81.2}\\
            
            \bottomrule
        \end{tabular}
        }
        \caption{ShapeNet classification accuracy with limited labeled training data for fine-tuning.\vspace{-1mm}}
        \label{table-shapenet_classification_accuracy_retrain}
    \end{center}
    \vspace{-3mm}
\end{table}

\subsection{Ablation Study}
\label{subsec:ablation_study}

\begin{table}[t]
    \begin{center}
    \ra{1.0}
        \resizebox{0.45\columnwidth}{!}{%
        \begin{tabular}{@{} lrr @{}}
            \toprule
            \textbf{Dataset}                       & \textbf{FR-SR}        & \textbf{SR-FR}     \\
            \midrule
            DFAUST                                 & 27.7                  & 4.4                \\
            3DMatch                                & \textbf{4.1}          & \textbf{3.8}       \\
            \bottomrule
        \end{tabular}
        }
        \caption{Results of using features pre-trained on different datasets in the downstream task of unsupervised non-rigid shape matching. 
        Values are mean geodesic error $\times 100$ on unit-area shapes.}
        \label{tab:dataset_ablation}
    \end{center}
    \vspace{-4.5mm}
\end{table}

We also evaluate the role of the pretraining dataset for local feature learning. 
For this, we compare 3DMatch used in our experiments to DFAUST \cite{dfaust:CVPR:2017}, a large-scale dataset of human subjects in motion.
We use the unsupervised shape matching task and the evaluation protocol introduced in \cref{subsec:human_matching} to test the generalization of a pre-trained feature extractor to the FR and SR datasets.


The comparisons in \cref{tab:dataset_ablation} show that pre-training on 3DMatch leads to more generalizable features and consistent matching performance, even though DFAUST has greater similarity to the downstream human shape datasets FR and SR.
We attribute this to the fact that \textit{local geometries }in 3DMatch, which consists of real-world scans, are richer and more complex than those in template-fitted DFAUST, leading to a more universally useful pre-training signal.

\begin{figure}[!t]
    \centering
    \includegraphics[width=\columnwidth]{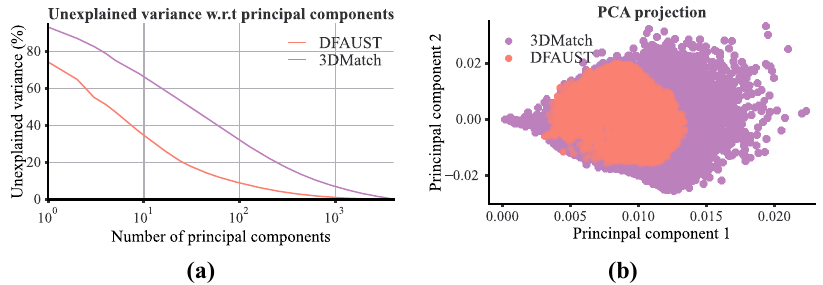}
    \caption{Comparing richness of local geometries in 3DMatch and DFAUST via PCA. (a) We perform PCA on sampled 3D local patches and plot the unexplained variance \wrt the number of principal components. (b) We project the local geometries onto the first two principal components.}
    \label{fig:ds_abla}
    \vspace{\figmargin}
\end{figure}

To validate this, we perform PCA analysis \cite{pca01} on the local patches of 3DMatch and DFAUST.
For each dataset, we first randomly extract 200K local patches. We then encode each patch as a high dimensional vector by first orienting it using a local reference frame and then voxelizing it to a small 3D grid of resolution = $16^3$ using the method of \cite{gojcic2019perfect}. The resulting vectors are 4096-dimensional and are fed as input to PCA. In \cref{fig:ds_abla} (a), we report the unexplained variance as a function of the number of principal components. It can be seen that 3DMatch is significantly more diverse than DFAUST since more principal components are needed to explain its full variance. In \cref{fig:ds_abla} (b), we visualize the projection of patches in the first two principal components and observe that local patches in DFAUST (red dots) are included in 3DMatch (blue dots), demonstrating the diversity and richness of 3DMatch once more.

\section{Conclusion and Limitations}
\label{sec:conclusion}
In conclusion, we have shown that our method of pre-training local features on rigid 3D scenes can generalize well to new and unseen classes of deformable organic shapes, enabling effective performance in various shape analysis tasks. Our study has highlighted the importance of selecting the right receptive field size to ensure feature transferability, which has led to the \textit{first general-purpose local feature pre-training}  for deformable shape analysis tasks. This research also sheds light on the relationship between rigid and non-rigid processing tasks, providing a link between two fields that have traditionally used different tools.

One limitation of our method is its reliance on differentiable voxelization, which can be memory and time-consuming, particularly during pre-training. Nonetheless, our results outperform PointContrast \cite{xie2020pointcontrast}, a point-based method that requires \textit{more training data} and has limited generalizability. Another limitation is that our features rely on LRF estimation, which might lack robustness to thin structures or boundaries of partial shapes. Exploring alternative scalable and robust local feature pre-training strategies is an fascinating direction for future work.

\mypara{Acknowledgements}
The authors would like to thank the anonymous reviewers for their valuable suggestions. 
Parts of this work were supported by the ERC Starting Grant No. 758800 (EXPROTEA) and the ANR AI Chair AIGRETTE.

\newpage




\twocolumn[{%
 \centering
 {\Large \bf Supplementary Materials for:\\Generalizable Local Feature Pre-training for Deformable Shape Analysis \par}
 {\vspace*{24pt}}
      {
      \large
      \lineskip .5em
      \begin{tabular}[t]{c}
        Souhaib Attaiki \hspace{1.5cm} Lei Li \hspace{1.5cm} Maks Ovsjanikov\\
LIX, \'Ecole Polytechnique, IP Paris
      \end{tabular}
      \par
      }
      \vskip .5em
      \vspace*{12pt}
}]
\appendix

In this document, we collect all the results and discussions, which, due to the page limit, could not find space in the main manuscript.
This supplementary material consists of two parts.
First, in \cref{suppsec:implementation_details}, we describe more implementation details mainly regarding our pilot study, local feature pre-training, and experiments on downstream deformable shape data.
Next, in \cref{suppsec:additional_results}, we present additional experimental results and analysis of our local feature pre-training strategy and its generalization in downstream tasks, including deformable shape matching and segmentation.

\section{Implementation Details}
\label{suppsec:implementation_details}

\subsection{Feature Locality vs. Transferability}
\label{suppsubsec:feature_locality_vs_transferability}
In Sec.~3 of the main text, we conducted a pilot study on feature locality vs. transferability on deformable shapes.
We tested three different architectures for pre-training a \textit{local} feature extractor, and their details are as follows.

\mypara{SparseConv.}
We used the \texttt{ResNet14} architecture introduced in \cite{choy20194d}.
During pre-training, given a 3D point cloud $P$, a fixed-size local patch with a radius of 0.15 is cropped at point $\mathbf{p} \in P$ and then reoriented with a local reference frame (LRF) computed by the method in \cite{gojcic2019perfect} for rotation invariance.
The resulting local patch is fed to the sparse convolution network, which extracts a 32-dimensional feature vector for point $\mathbf{p}$.

\mypara{PCPNet.}
It is a variant of PointNet \cite{qi2017pointnet} endowed with a quaternion spatial transformer.
We used the single-scale architecture proposed by \cite{pcpnet2018}.
PCPNet is designed to be a local network requiring input patches to have a fixed number of points.
Thus during pre-training, a fixed-size local patch (radius = 0.15) is cropped at point $\mathbf{p}$ and reoriented by an LRF. 
The local patch is then resampled to 1,024 points and fed to the network, resulting in a 32-dimensional feature vector for point $\mathbf{p}$.

\mypara{3DCNN.}
We used the architecture from \cite{li2021updesc} with a learnable receptive field size and differentiable voxelization, the same as our \OurMethodName{} in Sec.~4.1 of the main text.
More details can be found in \cref{suppsubsec:local_feature_pretraining}.

\mypara{Dataset.}
We pre-trained the above local networks on the 3DMatch dataset, which is a collection of RGB-D scan datasets with 62 indoor scenes and 4,142 point cloud fragments. 
There are 13K points on average in a fragment after downsampling.

\mypara{Loss.}
We used the PointInfoNCE loss, in which 300 point correspondences were randomly sampled for a pair of point clouds for faster training and the temperature parameter $\tau$ was set to 0.07. 

We also used the cycle consistency loss $\mathcal{L}_c$. During pre-training, we use the extracted features to build correspondences for rigid alignment between shapes $P$ and $Q$.
The intuition for $\mathcal{L}_c$ is that the estimated transformation $(\mathbf{R}, \mathbf{t})$ aligning $P$ to $Q$ should be the inverse of the transformation $(\mathbf{R}', \mathbf{t}')$ aligning $Q$ to $P$.
Mathematically, this can be expressed as:

\begin{equation}
\begin{bmatrix}
\mathbf{R} & \mathbf{t}\\
\mathbf{0} & 1
\end{bmatrix}
\begin{bmatrix}
\mathbf{R'} & \mathbf{t'}\\
\mathbf{0} & 1 
\end{bmatrix}
=
\begin{bmatrix}
\mathbf{R}\mathbf{R'} & \mathbf{R}\mathbf{t'} + \mathbf{t}\\
\mathbf{0} & 1
\end{bmatrix}
= \mathbf{I}
\end{equation}

\mypara{Application to deformable shape matching.}
In Fig. 3 of the main text, we have shown the results of shape matching on the Faust Remeshed dataset, directly using the pre-trained feature extractors. Given two shapes $S_1$, and $S_2$, we compute their respective point-wise features $F_1$ and $F_2$ using a specific pre-trained model. We first produce an estimate of the point-to-point maps $T_{21}^{nn}$ and $T_{12}^{nn}$ using nearest neighbor search between $F_1$ and $F_2$. We then filter the correspondences by mutual check: a pair of points $x \in S_1, y \in S_2$ is considered to be in correspondence, if and only if in the feature space, $x$ is the nearest neighbor of $y$, and $y$ is the nearest neighbor of $x$. This results in two filtered maps $T_{21}^{mf}$ and $T_{12}^{mf}$. Finally, we further refine these two maps using the ZoomOut method \cite{Melzi_2019}, which is based on navigating between the spectral and spatial domains while progressively increasing the number of spectral basis functions. We emphasize that if the initial point-to-point map is noisy or contains strong ambiguities like symmetry ambiguities, ZoomOut is not able to remedy these errors, thus leading to final correspondences of bad quality. We perform 10 iterations of ZoomOut, starting from 30 eigenfunctions up to 100 eigenfunctions.

\subsection{Local Feature Pre-training}
\label{suppsubsec:local_feature_pretraining}
In Sec.~4.1 of the main text, we introduced our local feature pre-training strategy.

\mypara{Feature extraction.} We use $r_{\text{LRF}}=0.3$ and $\sigma=10^{-3}$ for differentiable voxelization \cite{li2021updesc}, and the voxel grid resolution is set to $16^3$.
We pre-trained on the 3DMatch dataset introduced in \cref{suppsubsec:feature_locality_vs_transferability}.

\mypara{Pre-training loss.} 
For the PointInfoNCE loss $\mathcal{L}_{\text{nce}}$, its settings are described in \cref{suppsubsec:feature_locality_vs_transferability}.
For the cycle consistency loss $\mathcal{L}_{\text{c}}$, 300 points were randomly sampled on each point cloud for feature extraction and alignment estimation.
A relaxation-based solver is used in $\mathcal{L}_{\text{c}}$ for estimating a 3D transformation between two point clouds, and its details can be found in \cite{li2021updesc}.
 
In the main text, we investigated the performance difference between the cycle consistency loss and PointInfoNCE loss w.r.t learned feature smoothness.
Suppose that $F \in \mathbb{R}^{m \times n}$ is the matrix of extracted $n$-dimensional point-wise features for a shape of $m$ vertices, we measure the Dirichlet energy as follows: 
\begin{equation}
    E_{Dirichlet}(F) = \frac{1}{n} \sum_{i=1}^n F_i^{\top} W F_i,
\end{equation}
where $F_i$ is the $i^{\text{th}}$ column of $F$, and $W$ is the standard stiffness matrix computed using the classical cotangent discretization scheme of the Laplace-Beltrami operator \cite{Pinkall1993}.

\subsection{Baselines}
In Sec.~5 of the main text, we tested our proposed \OurMethodName{} features against a wide spectrum of competitors, including both hand-crafted and learned features.

Specifically, the Heat Kernel Signature (HKS) and Wave Kernel Signature (WKS) features are both sampled at 100 values of energy \textit{t}, logarithmically spaced in the range proposed in their respective original papers.
SHOT descriptors are 352-dimensional, and we used the implementation from the PCL library \cite{Rusu_ICRA2011_PCL}.
PointContrast features are 32-dimensional, and we used the publicly available implementation and the pre-trained weights released by the authors\footnote{\url{https://github.com/facebookresearch/PointContrast}}.

\subsection{Downstream Shape Analysis Training}
In Sec.~5 of the main text, we used DiffusionNet on top of the baselines features and our \OurMethodName{} respectively, in both the shape matching and segmentation tasks. We employed the publicly available implementation of DiffusionNet released by the authors\footnote{\url{https://github.com/nmwsharp/diffusion-net}}.
Unless specified otherwise, in our experiments, we used four DiffusionNet blocks of width = 128. 
The DiffusionNet is trained by an ADAM optimizer \cite{kingma2017adam} with an initial learning rate of $10^{-3}$.

In Sec.~5.1 of the main text, we also used a point-wise MLP network on top of the baselines features and our \OurMethodName{} respectively for supervised shape matching.
For this, we use the same MLP architecture as in FMNet \cite{litany2017deep}. 
After computing the point features with the MLP, we use them to compute the predicted functional map $C_{pred}$ as in \cite{donati2020deep} and penalize its deviation from the ground-truth map $C_{gt}$ using the L2 loss: $L = \|C_{pred} - C_{gt}\|_2^2$.

\subsection{Computational Specifications}
All our experiments were executed using Pytorch \cite{NEURIPS2019_9015}, on a 64-bit machine, equipped with an Intel(R) Xeon(R) CPU E5-2630 v4 @ 2.20GHz and an RTX 2080 Ti Graphics Card.

In terms of computational time, pertaining our method takes about 12 hours on a single RTX 2080 Ti Graphics Card, in contrast to the 64 hours required for PointContrast. The receptive field optimization takes about 20 minutes per dataset. For feature extraction, our method takes 3 seconds to extract local features for a 5000-vertex shape, which is on par with other local features like SHOT~\cite{tombari2010unique}, but slower than PointContrast (0.1s). Finally, the forward pass using \OurMethodName{} takes the same time as for all baseline features, e.g., 0.2 seconds per iteration for the unsupervised shape-matching experiment in Sec 5.1 of the main text.

\section{Additional Results and Analysis}
\label{suppsec:additional_results}

\subsection{Size of the learned receptive field}
\begin{figure}[t]
  \centering
  \includegraphics[width=0.99\linewidth]{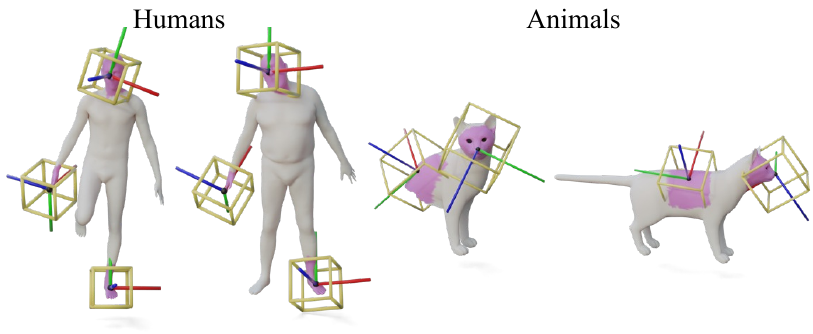}
   \caption{Visualizing the optimized receptive field for shape pairs.}
   \label{fig:receptive_field}
\end{figure}

Fig.~5 of our paper provides an illustration of the optimized receptive field in downstream tasks. In \cref{fig:receptive_field}, we include more visualizations for shape \textit{pairs} for both humans and animals.
Observe that the optimized receptive field indeed corresponds to interpretable concepts, such as the head or foot of a human, and is consistent across shape pairs.

\subsection{Human Shape Matching} 
\label{suppsubsec:human_matching}
In Sec.~5.1 of the main text, we performed unsupervised shape matching on the FAUST-Remeshed (FR), SCAPE-Remeshed (SR), and SHREC’19 datasets (SH) and reported the matching performance in Tab.~1.
We provide additional quantitative results of the FR-SR and SR-FR settings in \cref{tab:unaligned_unsup_supp}.
Compared with the baseline features, our \OurMethodName{} has the best and most consistent performance in both settings.

\begin{table}[t]
    \begin{center}
    \ra{1.0}
        \resizebox{0.8\columnwidth}{!}{%
            \begin{tabular}{@{} lrr @{}}
                \toprule
                \textbf{Method / Dataset}                & \textbf{FR}-\textbf{SR} & \textbf{SR}-\textbf{FR} \\
                \midrule
                SURFMNET                                 & 15.2                    & 9.5                     \\
                Cyclic FMaps                             & 23                      & 23.2                    \\
                WSupFMNet                                & 27.1                    & 14.2                    \\
                Deep Shells                              & 6.0                     & \textbf{3.4}            \\
                \midrule
                DiffusionNet - XYZ                       & 25.7                    & 8.4                     \\ 
                DiffusionNet - HKS                       & 7.9                     & 23                      \\ 
                DiffusionNet - WKS                       & 4.2                     & 24.1                    \\ 
                DiffusionNet - SHOT                      & 7.2                     & 4.1                     \\ 
                DiffusionNet - PCH                       & 11.4                    & 8.7                     \\ 
                DiffusionNet - PCN                       & 20.4                    & 9.1                     \\ 
                DiffusionNet - \OurMethodName{} (ours)   & \textbf{4.1}            & 3.9                     \\ 
                \bottomrule
            \end{tabular}
        }
        \caption{Accuracy of various features for unsupervised shape matching on un-aligned data.  X-Y means train on X and test on Y. Values are mean geodesic error $\times 100$ on unit-area shapes.}
        \label{tab:unaligned_unsup_supp}
    \end{center}
\end{table}



\subsection{Molecular Surface Segmentation} 
\label{suppsubsec:qualitative_evaluation}
\begin{figure}[t]
    \centering
    \includegraphics[width=\linewidth]{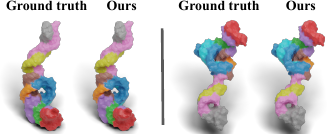}
    \caption{Qualitative evaluation of RNA segmentation on the dataset of \cite{poulenard2019effective}. Left: ground truth. Right: prediction by DiffusionNet + \OurMethodName{}.}
    \label{fig:rna_seg_qual}
\end{figure}

In \cref{fig:rna_seg_qual}, we show qualitative results of RNA segmentation using DiffusionNet + \OurMethodName{}.
It can be seen that the challenging RNA molecules can be robustly segmented into functional components with our pre-trained features.

\subsection{Human Shape Segmentation}
\label{suppsubsec:human_segmentation}

We performed an additional experiment on the human shape segmentation task.
We used the dataset introduced in \cite{maron2017convolutional}, which combines segmented human models taken from a variety of existing datasets.
We used the same train/test split of 380 training and 18 test shapes as in prior works.
We compared our \OurMethodName{} only with methods that used the original evaluation protocol as in \cite{maron2017convolutional}, i.e., without using post-processing and evaluating the results on the full shape resolution (techniques such as Mesh Walker \cite{lahav2020meshwalker} are thus excluded). 

\begin{figure}[t]
    \centering
    \includegraphics[width=\linewidth]{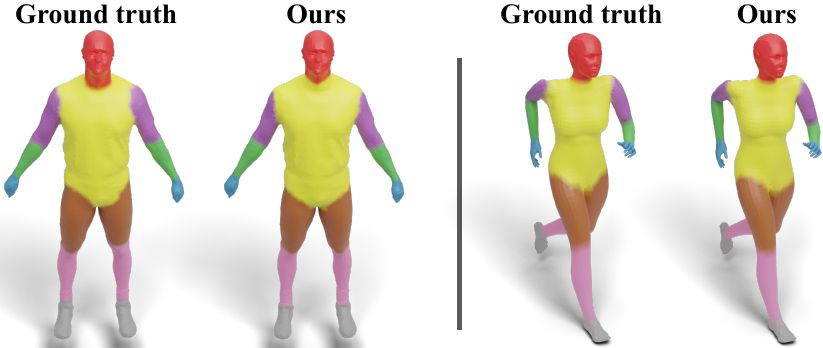}
    \caption{Qualitative evaluation of human shape segmentation on the dataset of \cite{maron2017convolutional}. Left: ground truth. Right: prediction by DiffusionNet + \OurMethodName{}.}
    \label{fig:human_seg_qual}
\end{figure}

We ran each experiment five times and report the mean and standard deviation of the accuracy in \cref{tab:human-segmentation}.  
Our \OurMethodName{} features achieve an accuracy of $92.4 \pm 0.25\%$, the state-of-the-art result on this dataset.
In \cref{fig:human_seg_qual}, we present qualitative results of human segmentation using DiffusionNet + \OurMethodName{}.
Note that the segmentation results are simply the network predictions, and we do not perform any complex post-processing to the segmentation.

\begin{table}[t]
    \begin{center}
    \ra{1.0}
        \begin{tabular}{@{}lr@{}}
            \toprule
            \textbf{Method}                           & \textbf{Accuracy $\pm$ s.d}                   \\
            \midrule
            GCNN \cite{masci2015geodesic}             & 86.4\%                                        \\
            ACNN \cite{boscaini2016learning}          & 83.7\%                                        \\
            Toric Cover \cite{maron2017convolutional} & 88.0\%                                        \\
            PointNet++ \cite{qi2017pointnet}          & 90.8\%                                        \\
            MDGCNN \cite{poulenard2019effective}      & 88.6\%                                        \\
            DGCNN \cite{wang2019dgcnn}                & 89.7\%                                        \\
            SNGC \cite{haim2019surface}               & 91.0\%                                        \\
            CGConv \cite{yang2021continuous}          & 89.9\%                                        \\
            \cmidrule{1-2}
            DiffusionNet - XYZ                        & 91.9 $\pm$ 0.27\%                             \\
            DiffusionNet - HKS                        & 91.5 $\pm$ 0.21\%                             \\
            DiffusionNet - WKS                        & 91.8 $\pm$ 0.33\%                             \\
            DiffusionNet - SHOT                       & 91.5 $\pm$ 0.77\%                             \\
            DiffusionNet - PCH                        & 85.6 $\pm$ 0.75\%                             \\
            DiffusionNet - PCN                        & 87.3 $\pm$ 0.57\%                             \\
            DiffusionNet - \OurMethodName{} (ours)    & \textbf{92.4 $\pm$ 0.25\%} \accuchange{+0.9}  \\
            \bottomrule
        \end{tabular}
        \caption{Human shape segmentation on the dataset of \cite{maron2017convolutional}. Our \OurMethodName{} achieves the state-of-the-art performance among methods that do not perform post-processing and evaluate on the full shape resolution. The reported numbers are the mean and standard deviation of the accuracy over five runs initialized randomly.}
        \label{tab:human-segmentation}
    \end{center}
\end{table}

\subsection{Robustness to Noise}
\begin{figure}[t]
    \centering
    \includegraphics[width=1.0\columnwidth]{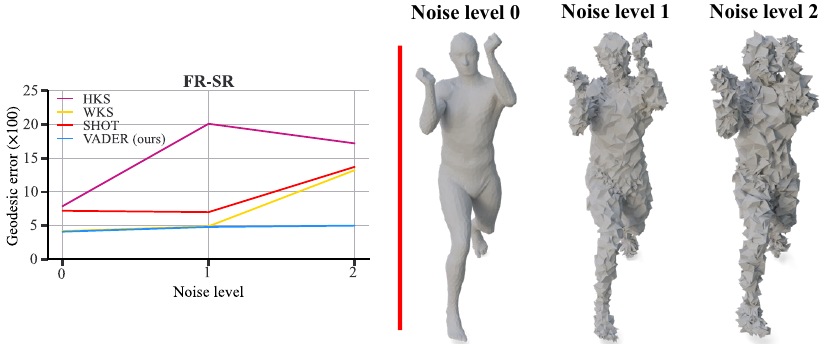}
    \caption{Left: Evolution of the geodetic error as a function of different
input noise levels. Right: Qualitative visualization of noise levels.}
    \label{fig:noise_robust}
\end{figure}

We performed an additional experiment to evaluate the robustness of our features to noise. For this, we followed the same setup as in Sec.~5.1 of the main text and in \cref{suppsubsec:human_matching}, by performing unsupervised learning on FR and testing on SR with an increasing amount of noise as input.
We compared our method to the best three competing features. The results are shown in \cref{fig:noise_robust} - left. It can be seen that our features are more robust to noise, i.e., the performance does not vary much with different noise levels (the intensity of the noise can be seen in \cref{fig:noise_robust} - right), which is not the case with other features, such as SHOT, whose performance degrades very quickly.

\subsection{Convergence Speed}

\begin{figure}[t]
    \centering
    \includegraphics[width=0.9\columnwidth]{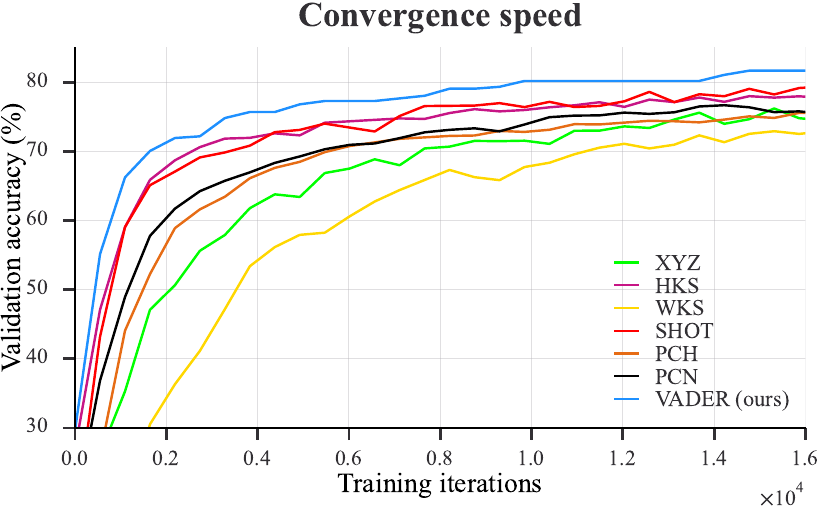}
    \caption{Evolution of the RNA segmentation accuracy on the validation set, during the training of DiffusionNet with different features.}
    \label{fig:convergence_speed}
\end{figure}

In our experiments, we observed that our \OurMethodName{} descriptors take less time to train and facilitate learning. To demonstrate this, we show in \cref{fig:convergence_speed} the evolution of validation accuracy during learning of the RNA segmentation task (Sec. 5.2 of the main text). It can be seen that compared to the other features, VADER requires far fewer training iterations to achieve similar performance. This clearly indicates the better descriptiveness and generalizability of our features.

{\small
    \bibliographystyle{ieee_fullname}
    \bibliography{references}
}

\end{document}